# Flower Categorization using Deep Convolutional Neural Networks

Ayesha Gurnani
L. D. College of Engineering
gurnani.ayesha.52@ldce.ac.in

Viraj Mavani
L. D. College of Engineering
mavani.viraj.604@ldce.ac.in

Vandit Gajjar
L. D. College of Engineering
gajjar.vandit.381@ldce.ac.in

Yash Khandhediya
L. D. College of Engineering
khandhediya.yash.364@ldce.ac.in

## Abstract

*We have developed a deep learning network for classification of different flowers. For this, we have used Visual Geometry Group's 102 category flower data-set having 8189 images of 102 categories from Oxford University. The method is basically divided in two parts i.e. Image segmentation and classification. We have compared two different Convolutional Neural Network architectures GoogleNet and AlexNet for the classification purpose. By keeping same hyper-parameters for both the architectures, we have found that the Top-1 and Top-5 accuracies of GoogleNet are 47.15% and 69.17% respectively whereas the Top-1 and Top-5 accuracies for AlexNet are 43.39% and 68.68% respectively. These results are extremely good when compared to random classification accuracy of 0.98%. This method for classification of flowers can be implemented in real-time applications and can be used to help botanists for their research as well as camping enthusiasts.*

## 1. Introduction

Flowers are everywhere around us. They can feed insects, birds, animals and humans. They are also used as medicines for humans and some animals. A good understanding of flowers is essential to help in identifying new or rare species when came across. This will help the medicinal industry to improve. The system proposed in the paper can be used by botanists, campers and doctors alike. This can be extended as an image search solution where photo can be taken as an input instead of text in order to get more information about the subject and search accordingly for best matching results.

As the classification of flower species is an important task, it is already in research and many different approaches have been developed. Previously, methods like Deformable Part Models [7], Histogram of Oriented Gradients [8] and Scale invariant feature transform [9] were used for feature extraction, linear classifiers and object detectors [10]. Later the work was focused on segmentation and classification using manual feature engineering. But nowadays, state-of-art performance is achieved by Convolutional Neural Networks. CNNs have fulfilled the demand of robustness and have removed the need of hand crafted features. They are similar to Artificial Neural networks but does not require feature engineering. Each neuron receives some inputs, performs a dot product and optionally follows it with a non-linear operation. At the last, CNNs also have a loss function which is to be minimized for optimization.

For using CNNs, a large amount of data is required for training. We have used the Visual Geometry Group's 102 category flower data-set used in [1] having 8,189 images spread over 102 categories from Oxford University. We split 15% of the total images for validation set and 15% for test set. Due to the large amount of data needed for CNNs, 8,189 images are not sufficient for training. Hence, we are using pre-trained models which are trained on ILSVRC2012 Dataset and fine-tuning them on the Oxford data-set. This makes the application less computational expensive. An example of the dataset is shown in figure 1.

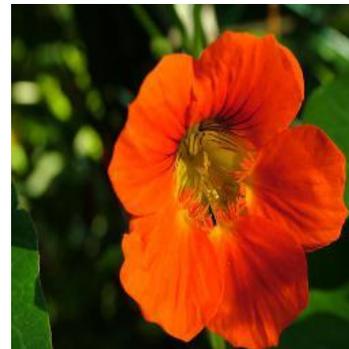

Figure 1. Image from Oxford data-set

In the given dataset [1], the images have too much data to be understood by the computer. The image needs to be segmented in order to extract the essential regions and remove the extraneous data. For this, we have used a segmentation technique [2] to segment the dataset. The data that we used has only the foreground of the images whereas the background has been removed by the above method. The segmented dataset is then given to the CNN as input and results have been observed. We compared the results of two different CNN architectures. Segmented image is shown in figure 2.

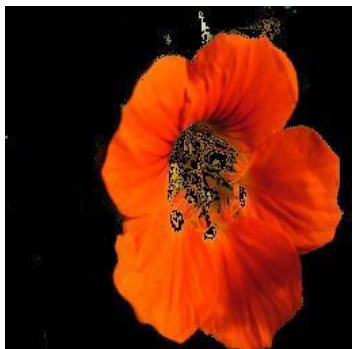

Figure 2. Segmented Image

## 2. Related Work

Not much work has been done to classify flowers as much is done to solve other problems in computer vision. Some of the previous works have used different feature based methods for flower classification like texture features and gray level co-matrix [11, 12]. Some newer works also include textual descriptions to assist deep convolutional neural networks for recognition [13].

Here, we have compared the performance of two different convolutional neural network architectures one of which is a legacy model and the other is the newest benchmark in the world of object detection and recognition. We have presented our findings in a comparison between the apparent performance of the models.

## 3. Proposed Method

For a given RGB image of any flower, our goal is to predict the category of the flower in the image. For this, first the image is segmented using a segmentation method [2].

The background will be removed and the image will have only the subject of interest with no background. These segmented images are then given to the CNN as input images for training. Therefore, the proposed method is divided into two main parts; segmentation and fine-tuning the deep convolutional neural network. For fine-tuning the CNN, we have used the ImageNet ILSVRC pre-trained models submitted to the competition.

### 3.1. Segmentation

We use the dataset of flower categories from Oxford University. The dataset consists of 8189 images having 102 categories of different flowers. The dataset is available here http://www.robots.ox.ac.uk/~vgg/data/flowers/102/. In figure 1, the diversity in background of different images can be seen. Therefore, it is necessary to remove the background from the images so that the subject of the image is highlighted. A segmentation method explained in [2] for segmenting the Oxford dataset is used. This segmented dataset contains images which have only the subject of the interest and the background is black.

The method basically finds out most frequently occurring hue values in the background and iteratively removes pixels with those values from the image. This eventually results in segmentation of the image as only the foreground remains at the end of the process.

### 3.2. Classification

#### 3.2.1 AlexNet

AlexNet as described in [3] is shown in figure 3 is one of the two models that we are using for classification. For training AlexNet, we have used NVIDIA DIGITS framework [4] with the BVLC Caffe [5] as backend. The training was performed on an NVIDIA TitanX GPU with the CUDA toolkit. For our experiments, we use pre-trained model of AlexNet trained on ILSVRC2012 dataset which has earlier been helpful in transfer learning problems. The AlexNet has . The network has a softmax layer at its output which showcases the likelihood of a particular class during inference.

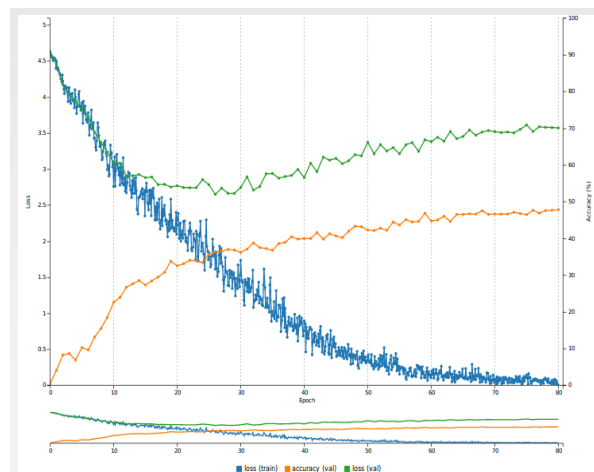

Figure 3. AlexNet Training Curve

For optimization, we used the stochastic gradient descent algorithm which computes the loss function in batches. The base learning rate used was the 0.01 with linear learning rate decay. We trained the network for 100 epochs. A dropout layer with a dropout ratio of 0.5 was also implemented to introduce regularization. At the end, we had a softmax layer to give out scores of each respective class. The network took 15 minutes 9 seconds to finish training. We observed the results on the test set and found the top-1 accuracy to be 43.39% and top-5 accuracy to be 68.68%. The training curve of the AlexNet can be found in figure 3.

*3.2.2 GoogleNet*

The GoogleNet is a convolutional neural network architecture implementing a deep module called the Inception as described in [6]. We fine-tuned this relatively newer model which was pre-trained on the ILSVRC2014 data-set for visual classification using the same system. The data-set fed to the network for fine-tuning was the same as that given to AlexNet. The data split also remains the same.

The GoogleNet was trained using a stochastic gradient descent algorithm which uses batch optimization. The base learning rate was 0.01 with linear learning rate decay. The network was trained for 100 epochs with dropout layer and softmax at output. The training on the NVIDIA TitanX GPU took 1hour 52minutes. The results observed were better than those of AlexNet but not to a great extent. The top-1 accuracy was 47.15% and top-5 accuracy was 69.17%. GoogleNet also offered higher generalizability because of superior regularization. The training curve of GoogleNet can be found in figure 4.

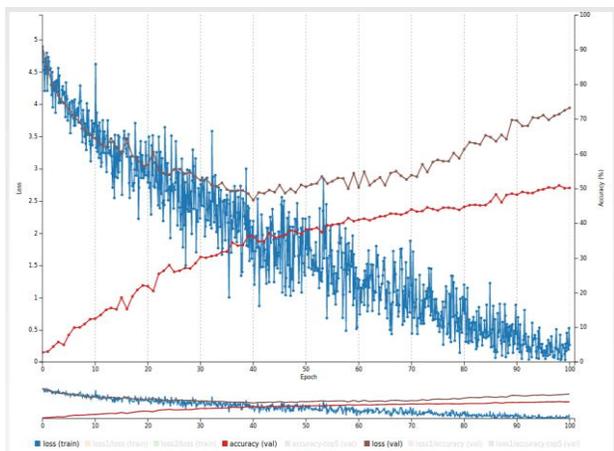

Figure 4. GoogleNet Training Curve

## 4. Comparison

Both the models provided similar results to an extent but still the relatively newer and advanced GoogleNet outperformed the AlexNet. For comparison, inference results on a single image for both the networks have been shown in figure 5.

Plus, there were some cases in which flowers wrongly classified by AlexNet but were picked up by GoogleNet. This was specifically observed in the class "Purple Cone-flower" which AlexNet confused with "Bougainvillea" while GoogleNet classified it correctly.

These results support two major claims by researchers worldwide: (i.) Deeper the network, better the performance and regularization on large datasets and (ii.) The reduction in number of parameters in the Inception module does not affect the accuracy of the model.

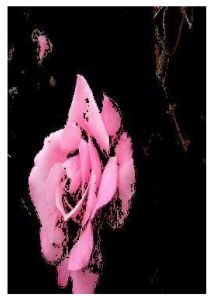

i. Prediction result from the AlexNet

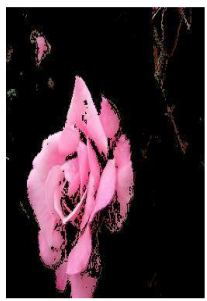

ii. Prediction result from the GoogleNet

Figure 5. Comparison in inference results

## 5. Conclusion

We segmented the data-set using the segmentation technique in [2] and gave the segmented data-set as the input to our CNNs. The test set and training set had 0% overlap and are completely different from each other. We trained the data on two CNN architectures i.e AlexNet and GoogleNet and have observed results for the same. We fine-tuned the models using the pre-trained weights of the architectures respectively.

We found that the Top-1 accuracy of AlexNet is 43.39% and Top-5 is 68.68%. When experimenting with GoogleNet, the Top-1 accuracy is 47.15% and Top-5 is 69.17% which is greater than AlexNet. Hence, GoogleNet offered better results in comparison with AlexNet because of various architectural differences. This model can be used for real-time applications to recognize flowers in the wild. Further research can also be done to improve the performance.